\documentclass[12pt]{article}
\usepackage{amsmath, amssymb, amsthm, float, graphicx}
\usepackage{xcolor}
\usepackage{parskip}
\definecolor{shadecolor}{rgb}{0.90,0.90,0.90}
\usepackage{cite}
\usepackage[linkbordercolor=blue]{hyperref}
\usepackage{subcaption}
\usepackage[utf8]{inputenc}
\usepackage[normalem]{ulem}
\usepackage[percent]{overpic}
\usepackage{mathtools}
\usepackage{setspace}

\numberwithin{equation}{section}

\hypersetup{colorlinks=false, linkcolor=blue, citecolor=red}

\def\beq{\begin{eqnarray}}\def\eeq{\end{eqnarray}}
\def\be{\begin{equation}}\def\ee{\end{equation}}

\def\v{\vec}
\def\pd{\partial}
\def\N{{\mathcal{N}}}

\def\s{\sigma}
\def\a{\alpha}
\begin{document}
\begin{titlepage}
\title{\bf Scaling and Resizing Symmetry in Feedforward Networks}
\date{}

\author{Carlos Cardona${}^{*}$,
  \\~~~~\\
  ${}^{*}$Physics Department, Case Western Reserve University,\\ Cleveland, OH 44106, USA
}

\maketitle
\vskip 2cm
\abstract{Weights initialization in deep neural networks have a strong impact on
  the speed of converge of the learning map. Recent studies have shown that in
  the case of random initializations, a
  chaos/order phase transition occur in the space of variances of random weights
  and biases. Experiments then had shown that large improvements can be made, in
  terms of the training speed, if a neural network is initialized on values
  along the critical line of such phase transition.  In this contribution, we
  show evidence that the scaling property exhibited by physical systems at
  criticality, is also present in untrained feedforward
  networks with random weights initialization at the critical line.
  Additionally,
  % by analogy between the covariance matrix propagation
  % through the network and the evolution towards equilibrium of a
  % two-dimensional statistical mechanical system
   we suggest an additional data-resizing symmetry, which is directly inherited from
  the scaling symmetry at criticality. }

\vfill {\footnotesize
	carlos.cardonagiraldo@case.edu}

\end{titlepage}
\newpage

%\tableofcontents

\onehalfspacing
%\doublespacing

\newpage
\section{Introduction}
Deep learning achievements during the last decade are quite impressive in
fields such as pattern recognition \cite{NIPS2012_c399862d}, speech recognition
\cite{hannun2014deep},  large language models \cite{radford2019language,
  brown2020language}, video and board games \cite{mnih2013playing,
  Silver2016MasteringTG} and neurobiology \cite{Yamins,McIntosh2017DeepLM}, just
to mention a few. This success has come with increasingly larger networks size, deep and complexity \cite{simonyan2015deep, szegedy2014going}, which often lead to undesired effects such a exploding or vanishing gradients \cite{279181} which hinders the minimization of the cost function.

Early work in deep learning \cite{279181, pmlr-v9-glorot10a} has shown that exploding or vanishing gradients crucially depend on weights initialization, and henceforth a properly chosen random distribution of weights at initialization can prevent such exploding or vanishing gradients.

The dependence of correlations on the deep of untrained random initialized feedforward networks has been
studied in \cite{Sompolinsky1988ChaosIR, poole2016exponential,
  schoenholz2017deep}, where it has been found that random Gaussian initialization develops
an order/chaos phase transition in the space of weights-biases variances, which
in turn translates into the development of vanishing/exploding gradients. Even
more interestingly, the existence of a critical line separating the ordered from
the chaotic phase, acts as a region where information can be propagated over
very deep scales, and as such, signals the ability to train extremely
deep networks at criticality. This phase transition has been observed so far also in
convolutional networks \cite{pmlr-v80-xiao18a}, autoencoders
\cite{DBLP:conf/iclr/LeNLPMVQ19} and recurrent networks \cite{pmlr-v80-chen18i}.

In this paper, we build on the results of \cite{poole2016exponential,
  schoenholz2017deep}, by looking a bit closer into the phase transition properties. After a quick review of the relevant results from
\cite{poole2016exponential, schoenholz2017deep}, we offer an analogous view for
the covariance matrix propagation through a feedforward network in terms of a
two-dimensional statistical physical system. Based on such analogy, we propose an order parameter associated to the phase transition and show that a scaling symmetry in
deepness arise at the critical phase, for which we compute numerically some
critical exponents. Then we argue at the level of conjecture, that such scaling
symmetry translates into a re-sizing symmetry for other dimension of the
feedforward network, such as input data size, hidden layers width and stochastic
gradient descent batch size. Lately, we perform experiments showing that
resizing down by a half the input data, hidden layers width or stochastic
gradient descent batch size has little detrimental impact on learning performance at the critical phase.

This results suggest that random initialization of feedforward networks at the
critical phase, might allow to train  networks with much less data or smaller
architectures, leading consequently to speed up learning. 

\section{Theoretical Background}\label{sec:The}
In this section we present few very well known definitions and properties of feedforward deep neural networks and set some notation.

\subsection{Feedforward Networks}
Let us start by considering a fully-connected untrained deep feedforward neural network with $L-$layers, each of width $N_{\ell}$, with $\ell$ denoting the corresponding layer label. Let $W^{\ell}$ be the matrix of weights linking the $\ell-1$ layer to the $\ell$ layer (hence of dimension $\dim(W^{\ell})=N_{\ell}\times N_{\ell-1}$) and $\v{b}_{\ell}$ the bias vector at layer $\ell$. The vector output at each layer (or post-activation) is then defined by the recursion relation:
\be
\v{x}^{\ell}=\phi(\v{h}^{\ell}) \,,\quad \v{h}^{\ell}=W^{\ell}\v{x}^{\ell-1}+\v{b}^{\ell}\,, 
\ee
with the initial condition $\v{x}^{0}= $input-data, of dimension $N_{0}$. The
activation function $\phi$ is a scalar function acting element-wise on the
vector components of its argument. For now we consider $\phi$ to have a sigmoid
shape \be\phi(\pm\infty)=\pm 1,\quad\phi(-x)=\phi(x) \,,\ee but otherwise to be
arbitrary \footnote{However, for the experiments later in this note we will use
  $\phi(x)=\tanh(x)$}.

The goal of the network is to learn the mapping,
\be
\v{x}^L=\Omega(\v{x}^0)\,,
\ee
that both, best approximate the input data to the output data, and that best generalizes to new data. This is done by minimizing a cost function measuring how far off the learned mapping is from the actual input-output map. Such cost function is in most cases taken to be the {\it sum of square errors}:
\be
C={1\over 2}\sum_{i=1}^{N_0}|\Omega(\v{x}_i^0)-\v{x}_i^{({\rm output\,\,data)}}|^2\,,
\ee
An important object characterizing $\Omega$ is the input-output Jacobian $J\in \mathbb{R}^{N_0\times N_L}$ given by:
\be
J={\pd \v{x}^{L}\over \pd \v{x}^{0}}=\prod_{\ell=1}^L D^{\ell}W^{\ell}\,,\ee
with the components of $D^{\ell}$ given by $D^{\ell}_{ij}=\phi'(\v{x}^{\ell})\delta_{ij}$.

We want to consider an initial value for the weights and biases to be drawn from the following normal distributions:  $W_{i,j}^{\ell}\sim\N(0,{\s_w^2\over N_{\ell-1}})$,  $b^{\ell}_i\sim\N(0,\s_b^2)$. The weight re-scaling over $N_{\ell-1}$ is to preserve the weights finite for large sizes of hidden layers. 
%% We additionally will consider an additional constrains on the matrix of weights ensembles: one in which the following orthogonality relation is satisfied:
%% \be
%% W^{\ell}(W^{\ell})^T=\s_w^2\mathbb{I}\,,
%% \ee
%% to which we will refer as the orthogonal ensemble.
\subsection{Mean Field Theory of Deep Feedforward Signal Propagation}
Our theoretical starting point follows from the results of
\cite{poole2016exponential, schoenholz2017deep}, which we now proceed to quickly
review, but with a slight physical spinoff.
\subsection{Physical Statistical System}
We want to understand how the
empirical distribution of pre-activations $\v{h}^{\ell}$ propagates through the
network. To do that, we can define the following partition function,
\be
Z=\sum_{E(\ell)} e^{-\beta {\cal H}_{\ell}} 
\ee
with a quadratic Hamiltonian $\cal{H}$ given by,
\be
{\cal H}^{(\ell)}=B\sum_{a,b}q^{\ell}_{a,b}\,,
\ee
with $B$ an arbitrary parameter and  $q^{\ell}_{a,b}$ being the covariance matrix,
\be
q^{\ell}_{a,b}={1\over N_{\ell}}\sum_{i=1}^{N_{\ell}}h_i^{\ell}(\v{x}^{0,a})h_i^{\ell}(\v{x}^{0,b}) \,,\quad a,b=1,\cdots, N_0\,.
\ee
We additionally define the {\it free energy} ${\cal F}$ as it's usual in physical statistical systems,
\be
{\cal F}=-{1\over \beta}\log Z\,,
\ee
from which we can see the following relation
%\footnote{Here is assume that the mean value is the same for all pairs $(a,b)$},
\be
{\pd {\cal F} \over \pd B}={1\over Z}\sum_{a,b}\sum_{E(\ell)}q^{\ell}_{a,b}\,
e^{-\beta{\cal H_{\ell}}}\equiv
\sum_{a,b}\langle  q^{\ell}_{a,b} \rangle\,.
\ee
As we will see in next subsection, as $\langle q^{\ell}_{a,b} \rangle$
propagates through the network, after few layers it will eventually reach an
equilibrium constant value $q^*_{a,b}$, i,e,
\be
{\pd {\cal F} \over \pd B}=\sum_{a,b} q_{a,b}^*\,.
\ee
For convenience, let us split the Hamiltonian in diagonal and non-diagonal
parts,
\be
{\cal H}^{(\ell)}={\cal H}^{(\ell)}_d + {\cal H}^{(\ell)}_{nd}\,,
\ee
with
\be\label{qmap2}
{\cal H}^{(\ell)}_d=B_d\sum_{a}q^{\ell}_{a,a}\,,\quad {\cal H}^{(\ell)}_{nd}=B_{nd}\sum_{a\neq b}q^{\ell}_{a,b}\,.\ee

For large $N_{\ell}$, it has been shown in \cite{poole2016exponential,
  schoenholz2017deep} that $\v{h}^{\ell}$ converge to a zero mean Gaussian
distribution over the given layer $\ell$, and in such limit, we can replace the
sums for integrals with the corresponding Gaussian probability density (see
\cite{poole2016exponential} for details). At equilibrium, we can write  the
diagonal Hamiltonian as,
\be
{\cal H}^*_d=B_d\sum_{a}\left[\s_{w}^2\int_{-\infty}^{\infty}{dz\over
  \sqrt{2\pi}}\,e^{-{z^2\over 2}}\,\phi(\sqrt{q_{a,a}^{*}}z)^2 +\s_b^2\right]\,,
\ee
whereas the non-diagonal looks like,
\be
{\cal H}^*_{nd}=B_{nd}\sum_{a\neq b}\left[\s_{w}^2\int_{-\infty}^{\infty}{dz_a dz_b\over 2\pi}\,e^{-(z_a^2+z_b^2)\over 2}\,\phi(\sqrt{q_{a,a}^{*}}z_a)\,\phi(u_b) + \s_b^2\right]\,,
\ee
with,
\be 
u_b=\sqrt{q_{b,b}^{*}}\left[c^{*}_{a,b}z_1+\sqrt{1-(c^{*}_{a,b})^2}\,z_2\right]\,,\quad c^*_{a,b}={q_{a,b}^*\over \sqrt{q_{a,a}^{*}}\sqrt{q_{b,b}^{*}}}\,,
\ee
here $z_a$ and $z_b$ are independent standard Gaussian variables, while $u_a$
and $u_b$ are correlated Gaussian variables with covariance matrix
$q_{a,b}^{\ell-1}$, which become $q_{a,b}^{*}$ at equilibrium. 

\begin{figure}[h]
  \includegraphics[width=0.52\textwidth]{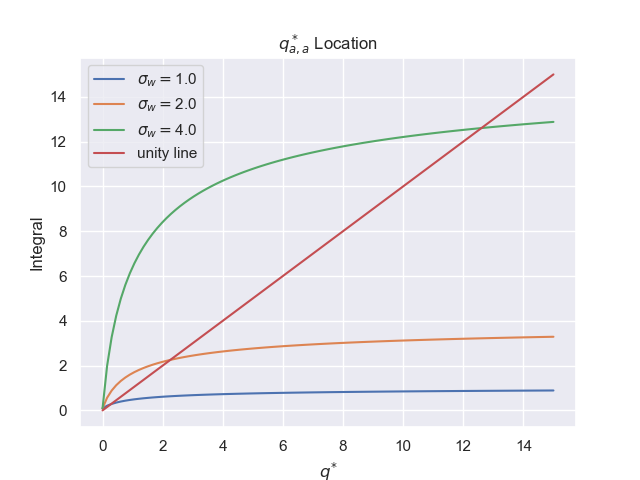}  \includegraphics[width=0.52\textwidth]{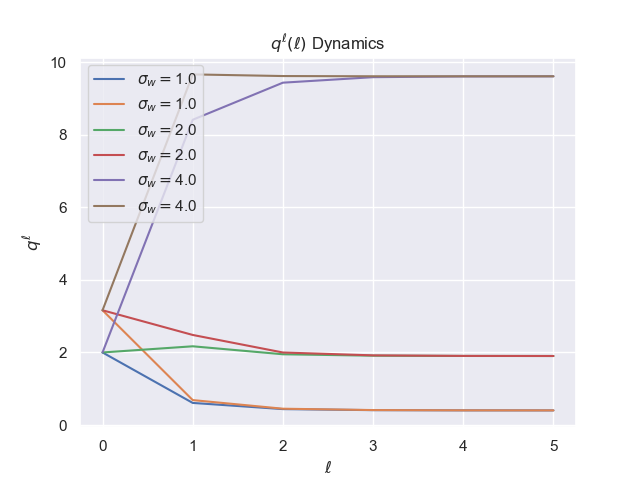}
  \caption{Left: Intersection of unity line with integral at RHS of \eqref{fd}
    for three different $\sigma_w$ and $\sigma_b=0.3$. Right: Dynamics of
    $q^{\ell}$ across layer propagation, for two different values of the initial
    condition. To make both plots, we have used $\phi(z)={\rm tanh}(z)$}
  \label{fig1}
\end{figure}

\begin{figure}[h]
  \includegraphics[width=0.52\textwidth]{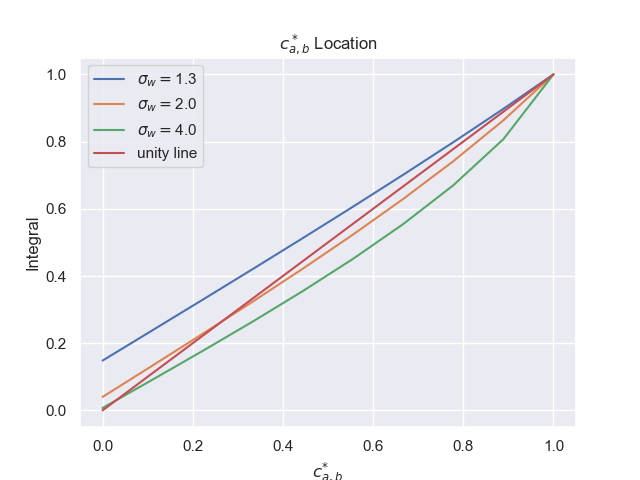}  \includegraphics[width=0.52\textwidth]{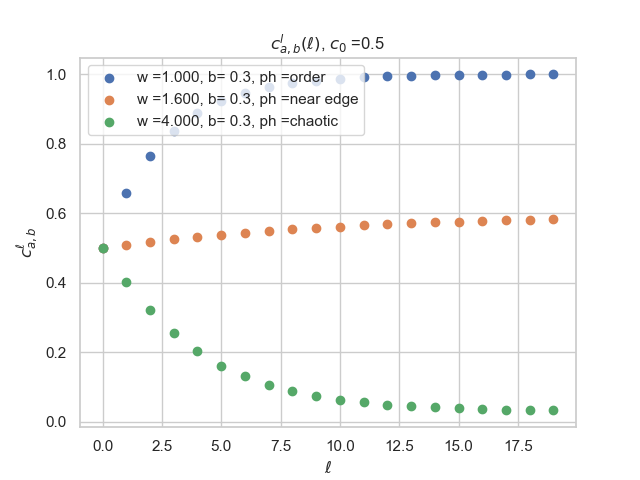}
  \caption{Left:  Intersection of unity line with integral at RHS of \eqref{fnd}
    for three different $\sigma_w$ and $\sigma_b=0.3$. Right: Dynamics of
    $c^{\ell}$ across layer propagation, for two different values of the initial
    condition. To make both plots, we have used $\phi(z)={\rm tanh}(z)$} 
  \label{fig2}
\end{figure}

Once the system reach equilibrium, we approximate the
partition function by,
\be
Z_d=e^{-\beta\, L\,{\cal H}_d^*-\beta\, L{\cal H}_{nd}^*}\,,
\ee
where $L$ is the total number of layers. From it, we have
\be
{\cal F}=L{\cal H}_d^*+L{\cal H}_{nd}^*\,,
\ee
which lead us to the following consistency equation,
\be\label{fd}
{\pd {\cal F} \over \pd B_d}=q_{a,a}^*=\s_{w}^2\int_{-\infty}^{\infty}{dz\over
  \sqrt{2\pi}}\,e^{-{z^2\over 2}}\,\phi(\sqrt{q_{a,a}^{*}}z)^2 + \s_b^2\,,
\ee
with a similar equation for the non-diagonal,
\be\label{fnd}
{\pd{\cal F}\over B_{nd}}=q_{a,b}^*=\,\s_{w}^2\int_{-\infty}^{\infty}{dz_a dz_b\over 2\pi}\,e^{-(z_a^2+z_b^2)\over 2}\,\phi(\sqrt{q_{a,a}^{*}}z_a)\,\phi(u_b) +\s_b^2\,.
\ee
The solutions to these consistency conditions, or fixed points, correspond to the equilibrium
values of the covariance matrix. They can be solved very efficiently numerically,
by pinning down the interception of the unity line with the value from the
integrals at the right hand side of both conditions, as illustrated at the left
frames of figures 
\ref{fig1} and \ref{fig2},

\subsection{Dynamical System}
Although the storytelling in subsection above might be useful to build some physical
intuition, it does not provides us with a good view of the non-equilibrium dynamics that the system
undergoes. For that it will convenient to resort back to  \cite{poole2016exponential,
  schoenholz2017deep}, where instead of focusing solely on
the equilibrium state, the random network has been approached from the point of view of
dynamical systems, by considering the recursion relations leading to \eqref{fd}
and \eqref{fnd},
\be\label{qmap}
q^{\ell}_{a,a}=\s_{w}^2\int_{-\infty}^{\infty}{dz\over \sqrt{2\pi}}\,e^{-{z^2\over 2}}\,\phi(\sqrt{q_{a,a}^{\ell-1}}z)^2 + \s_b^2\,,
\ee

\be
q^{\ell}_{a,b}=\s_{w}^2\int_{-\infty}^{\infty}{dz_a dz_b\over 2\pi}\,e^{-(z_a^2+z_b^2)\over 2}\,\phi(\sqrt{q_{a,a}^{\ell-1}}z_a)\,\phi(u_b) + \s_b^2\,,
\ee
with,
\be
u_b=\sqrt{q_{b,b}^{\ell-1}}\left[c^{\ell-1}_{a,b}z_1+\sqrt{1-(c^{\ell}_{a,b})^2}\,z_2\right]\,,\quad c_{a,b}={q_{a,b}\over \sqrt{q_{a,a}^{\ell-1}}\sqrt{q_{b,b}^{\ell-1}}}\,,
\ee
for the diagonal and non-diagonal components respectively. This recursions define
an {\it  iterative map}.
% The recursion relation for the diagonal component of the covariance matrix $q_{a,a}^{\ell}(q_{a,a}^{\ell-1})$ is an increasing monotonic concave function, which intersects the $q_{a,a}^{\ell}=q_{a,a}^{\ell-1}$ unity line, or in other words, has a fixed point $q_{a,a}^*$, as is shown in the left figure below for the diagonal component.

By iterating this recursion a few times, we can see the dynamics of the
covariance matrix before reaching equilibrium, as shown at the right frame in
figure \ref{fig1}. Notice that the fixed point $q_{a,a}^*$ is reached after few
layers.

For the non-diagonal component, it is convenient to use the normalized correlation coefficient $c_{a,b}$, to write instead,
\be\label{cmap}
c^{\ell}_{a,b}={\s_{w}^2\over q_{a,a}^*}\int_{-\infty}^{\infty}{dz_a dz_b\over 2\pi}\,e^{-(z_a^2+z_b^2)\over 2}\,\phi(\sqrt{q_{a,a}^*}z_a)\,\phi(u_b) + \s_b^2\,,
\ee
with
\be 
u_b=\sqrt{q_{a,a}^*}\left[c^{\ell-1}_{a,b}z_1+\sqrt{1-(c^{\ell}_{a,b})^2}\,z_2\right]\,.
\ee
$c_{a,b}$ exhibit a more interesting behavior: for low values of $\sigma_w$ it only has one fixed point at $c_{a,b}^*=1$ which is also stable, but for values larger than a given $\s_w^{\rm critical}$ ( $\s_w^{\rm critical}\sim 1.39$ for the cases depicted in figure \ref{fig2}), the recursion develops a second fixed stable point, as shown in the left figure \ref{fig2}, and the fixed point $c_{a,b}^*=1$ becomes unstable, signaling a phase transition in $(\s_w, \s_b)$ space.
For a given tuple of values $(\s_w, \s_b)$, we can check the stability of the covariance map by looking into the derivative of the iterative map at the fixed point,
\be
\chi_1\equiv{\partial c_{a,b}^{\ell}\over c_{a,b}^{\ell-1}}\Bigg|_{c^*}=\s_w^2\int_{-\infty}^{\infty}{dz\over \sqrt{2\pi}}\,e^{-{z^2\over 2}}\,\left(\phi'(\sqrt{q_{a,a}^*}z)\right)^2 \,,
\ee
so the critical values are those that solve the equation $\chi_1=1$.

From the right figure \ref{fig2}, we can see that it takes longer for the
$c_{a,b}^*$ fixed point to be reached than it takes to reach $q_{a,a}^*$. In figure \ref{fig2} we
have used $\s_b=0.3$ in all cases, in which case the critical $\s_w$ turns out to
be $\s_w^{\rm critical}=1.39$. For values below $\s_w^{\rm critical}$ the
network correlates initial data, for values above $\s_w^{\rm critical}$ the
network uncorrelates initial data, and for values near $\s_w^{\rm critical}$ the
network tend to preserve the initial data correlations.
% From the
% point of view of dynamical systems, equations \eqref{qmap} and \eqref{cmap}
% define an iterative map, for which the authors of \cite{poole2016exponential,
%   schoenholz2017deep} have interpreted the correlation/uncorrelation behavior as an {\it order/chaos} phase transition
A property of a dynamical system signaling chaotic behavior is that by evolving two infinitesimally close initial conditions, they both reach values significatively apart from each other. Considering the covariance matrix, by initializing a network with random Gaussian weights at $\s_w>\s_w^{\rm critical}$, and tracking two input data points initially very correlated (very close in correlation space), they will get uncorrelated as they propagate through the network (separate from each other in correlation space), signaling a chaotic phase. Conversely, uncorrelated initial data points, will correlated for an initialization with  $\s_w<\s_w^{\rm critical}$, signaling an ordered phase. This two phases are related to exploding or vanishing gradients during stochastic gradient descent.

In the next section, we would like to take a closer look into the phase.

\section{Phase Transition}

\begin{figure}[h]
  \centering
  \includegraphics[width=0.7\textwidth]{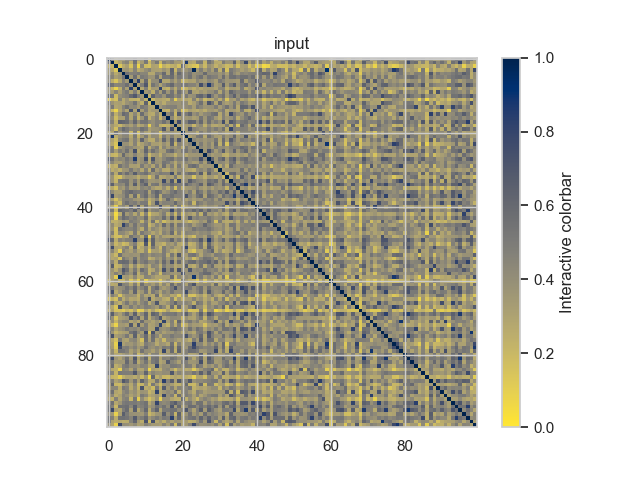} 
  \caption{Covariance matrix for 100 input data points from MNIST}
  \label{fig3}
\end{figure}

For the examples and experiments treated in the rest of this note, we will take the MNIST dataset
\cite{726791, MNIST} of handwritten numbers. Let us then consider the normalized correlation coefficient matrix, or
covariance matrix of such input data (depicted in figure \ref
{fig3} for a subset of 100 input data points for visualization clarity) as it propagates through the
network. Think of this matrix as a  two-dimensional lattice statistical mechanical system at some initial state (out of equilibrium), where each
``pixel'' or component of the matrix represents a physical quantity ( such a
magnetic moment, for example) whose value is between
zero and one.  Now consider that the system evolution is the same as the one
undergone by the covariance matrix as it propagates through a
feedforward network.  A given initial $\s_b$ can be thought of as an
external field (such a magnetic or electric field), and a given $\s_w$ as a temperature.
\begin{figure}[h]
  \includegraphics[width=0.52\textwidth]{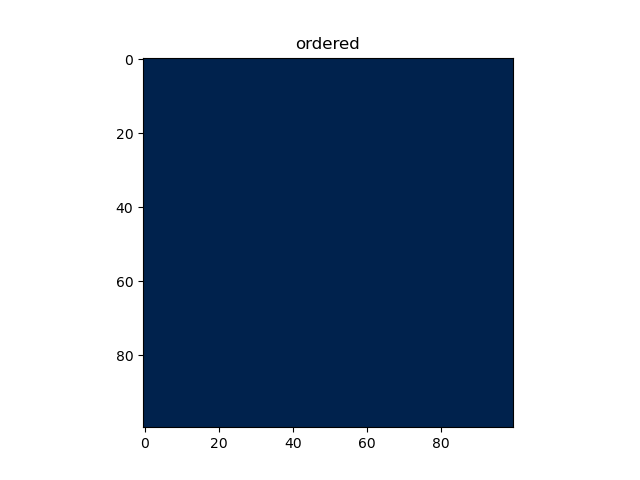}  \includegraphics[width=0.52\textwidth]{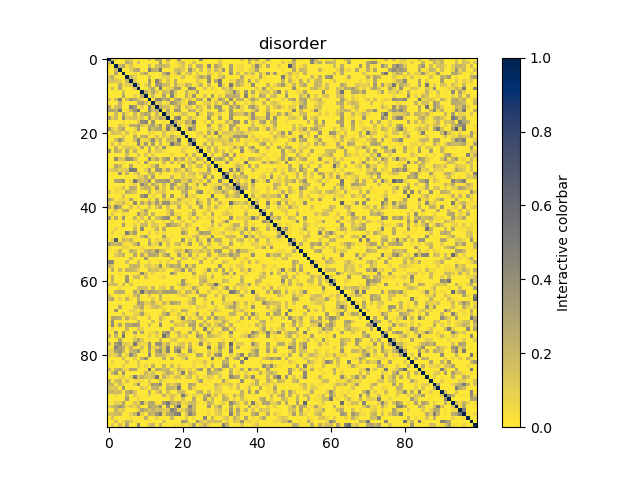} \begin{center} \includegraphics[width=0.52\textwidth]{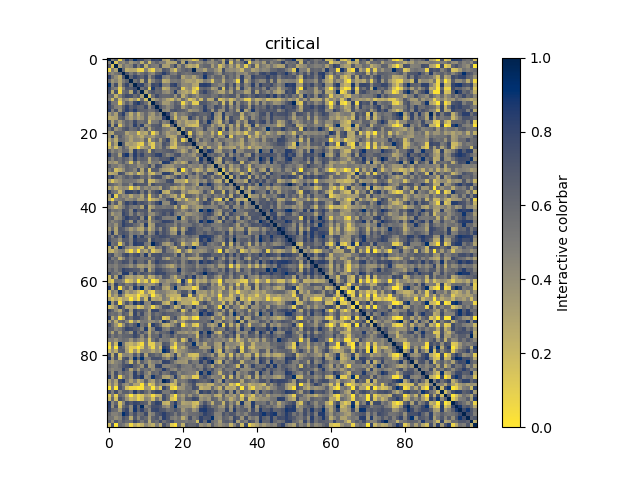}\end{center}
  \caption{Left: Correlation matrix output order phase, Right: Correlation matrix output disorder phase. Center Below: Correlation matrix output critical phase}
  \label{fig4}
\end{figure}

Now imagine that we set a ``temperature'' $\s_w$ and a ``magnetic field'' $\s_b$, and
let the system evolve towards equilibrium \footnote{$\s_w$ and $\s_b$ are
  only pictorial analogies of an actual temperature and external field, in
  particular they do not even have the proper units.}. Evolving in this context
means that we let the initial data propagate through the network with the given
random initialization. Once the data propagate across all layers in the order
phase, the covariance matrix looks like a system at zero temperature (completely
correlated),  the order phase looks like a system at infinite temperature (completely
uncorrelated) after propagation in the chaotic phase and a system at critical
temperature after propagating in the critical phase. To illustrate this, we
created a feedforward network with  10 hidden layers, each layer composed of 50 units \footnote{the output layer in
  this case, is not the one corresponding to the classification problem (ten
  digits layer), but an arbitrary 50 units layer.}
and propagate the MNIST data once over initializations corresponding to each of these three phases, then we compute the correlation matrix of the resulting output. The results are illustrated in the pictures at figure \ref{fig4}

By propagating the input data at the ordered phase, the network do a very good
job correlating the data whereas at the disorder phase the data gets almost
completely uncorrelated. At the critical phase however, even though the network
tends to correlated data slightly, it reach a fixed point where the data
propagate to a ``fix'' intermediate (non-saturated) correlation. Within the
analogy to statistical physics, we define the  following {\it order parameter}
\be
\langle c\rangle={2\over N}\sum_{a>b}c^*_{a,b}\,,
\ee
which we will call, the mean correlation. Here $N$ is the number of data points
in a given case. In the order phase, $\langle c \rangle$ is
close to one, and we have a  large mean correlation, at the disorder phase the
$\langle c \rangle$ is small and we have a vanishing mean correlation. Lastly, at the
critical phase, the mean correlation is in between the two other phases.
Henceforth, the change of  $\langle c \rangle$ from non-zero to zero signal a phase
transition, and that is why we have interpreted it as an order parameter.

Computing the {\it mean correlation} for the particular examples displayed in
figures \ref{fig3} and \ref{fig4} we get the values in table \ref{tab1}.
\begin{table}[ht]
  \caption{Mean Correlations}
\begin{center}
  \begin{tabular}{ |c|c|c|c|c| } 
    \hline
     & Input & Ordered & Disordered & Critical \\
    \hline
    $ \langle c\rangle$ & 0.414 & 0.99 & 0.041 & 0.50  \\
    \hline
  \end{tabular}
  \label{tab1} 
\end{center}
\end{table}
%  \CC{ try to push this analogy theoretically: For example, in the
  % usual monte carlo simulation for ising ferromagnet, we let the algorithm
  % evolve with a bolzamt probability distribution. Here, that evolution is given
  % by the propagation of the covariance matrix across the network, what is the
  % ralation between this two `markov-chains'? }
This values reflect the values of the corresponding fixed points.

An observation from the table above that will play a role in the next section's
discussion, is that the value of order parameter for the input data is
not too far from the value at the critical phase output.
\section{Scaling Symmetry}
It is well known that physical systems  having second order phase transition exhibit scaling symmetry at the critical phase. In this section, we ask the question if such symmetry is also exhibited in the  feedforward initializing network that we have discussed in the previous section.

\subsection{Theoretical Considerations}
One can ask how far the input data need to propagate in order for the
correlation matrix to reach its fix point. This question was detailed answered
in \cite{schoenholz2017deep}. For enough large $\ell$, the covariance
matrix approach the fix point exponentially,
\be\label{2pf}
|p_{a,b}^{\ell}-p^*|\sim e^{-{\ell \over \zeta_{p}}}\,,
\ee
with $p$ indicating that the propagation length $\zeta_{p}$ is different for the diagonal terms $p\equiv q$ than for the non-diagonals $p\equiv c$, respectively,
\be\label{za}
\zeta_{q}^{-1}=-\log\left[\chi_1+\s_w^2\int_{-\infty}^{\infty}{dz\over \sqrt{2\pi}}e^{-{z^2\over 2}}\phi''(z\sqrt{q^*})\phi(z\sqrt{q^*})\right]
\ee
and
\be\label{zc}
\zeta_{c}^{-1}=-\log\left[\s_w^2\int_{-\infty}^{\infty}{dz_a\over \sqrt{2\pi}}{dz_b\over \sqrt{2\pi}}e^{-{(z_a^2+z_b^2)\over 2}}\phi'(z_a\sqrt{q^*})\phi'(u_b^*)\right]\,.
\ee
$\zeta_p$ establish a depth scale that measures how far deep the input
information propagates into a random initialized network.

We are concerned about the correlation length at the fixed point, therefore
evaluating $\zeta_c$ at $c^*$, reduces it to \be \zeta_c=-{1\over\log\chi_1}\,.
\ee Since the phase transition occurs when $\chi_1=1$, the correlation length
diverges there, hence at the critical phase the information will propagate at
all deep lengths, as long as the fixed point has been already reached. This is
indeed a signal of scaling symmetry.

In experiments on physical systems as well as in solvable theories, it has been
observed that at the critical phase, when the correlation length diverges,
the behavior of the covariance (or two-point correlation function) given by \eqref{2pf}, breaks down, and  instead $p_{a,b}$
approaches the fixed point in a power-law fashion,
\be
|p_{a,b}^{\ell}-p^*|\sim {1\over \ell^{\a}}\,,
\ee
with $\a$ some coefficient, which in the physical literature corresponds to the
so-called {\it critical exponents}. For the current case of the feedforward
network, the break down of \eqref{2pf} near the phase transition can be seen
by realizing that its derivation at \cite{schoenholz2017deep} relies on a
perturbative expansion around $p^*$ with perturbation parameter given by
\be \epsilon^{\ell}\sim e^{-{\ell\over\zeta_p}}\,,\ee
which becomes of order one around the critical phase, hence breaking the
perturbative approximation.

We now want to check experimentally, if such power-law behavior from physical
statistical systems, is also observed in the random network phase transition discussed here.
\subsection{Critical Exponents for Information Propagation}
\begin{figure}[h]
  \centering 
  \includegraphics[width=0.7\textwidth]{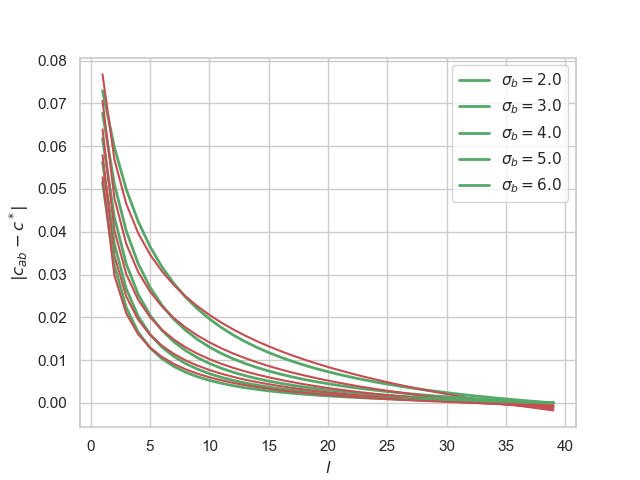}
  \caption{Power law behavior of information propagation at criticality. The red
  curves indicate the fit \eqref{powerlaw} for each $\sigma_b$}
  \label{fig4p}
\end{figure}

First we want to check that the power law behavior observed at criticality in
phase transitions of physical systems can be, at least approximately, observed
in the information propagation in random networks at the critical phase. In
order to see that, we have plotted $|c_{a,b}^{\ell}-c^*|$  for $\s_b=2, 3,
\cdots 6$ with an initial input $c_{a,b}$ very close to the critical value and fitted to each case a power law of the form,
\be\label{powerlaw}
|p_{a,b}^{\ell}-p^*|={c\over\ell^{\a}}+b\,,
\ee
the resulting plots and fits are shown in figure \ref{fig4p}, from where
we can see that indeed a power law behavior \ref{powerlaw} provides a good
approximation for the behavior of information propagation very close to
criticality. The correspondent fitted critical exponents for each $\s_b$ are shown in the table below,
\begin{table}[ht]
\begin{center}
  \begin{tabular}{ |c|c|c|c|c|c| } 
    \hline
   $\s_b$ & 2 & 3 & 4 & 5 & 6 \\
    \hline
  $\a$ & 0.213  & 0.406 & 0.545 & 0.653 & 0.743  \\
    \hline
 %   \label{tab2}
  \end{tabular}
\end{center}
\end{table}
\subsection{Exploring Scaling Symmetry in Covariance Space}

The divergence of the correlation length in layer space $\ell$ discussed above, provide us
with evidence for a scaling symmetry taking place at criticality. This scaling
symmetry tell us that, once the critical fixed point is reached, the correlation matrix
is preserved through the subsequent layers.
%The random network works as a kind of projector, projecting the input
%correlation matrix into the output ``screen'' or output layer,  where the
%fidelity of the projection depends crucially on the phase.

\begin{figure}[!htb]
  \includegraphics[width=0.5\textwidth]{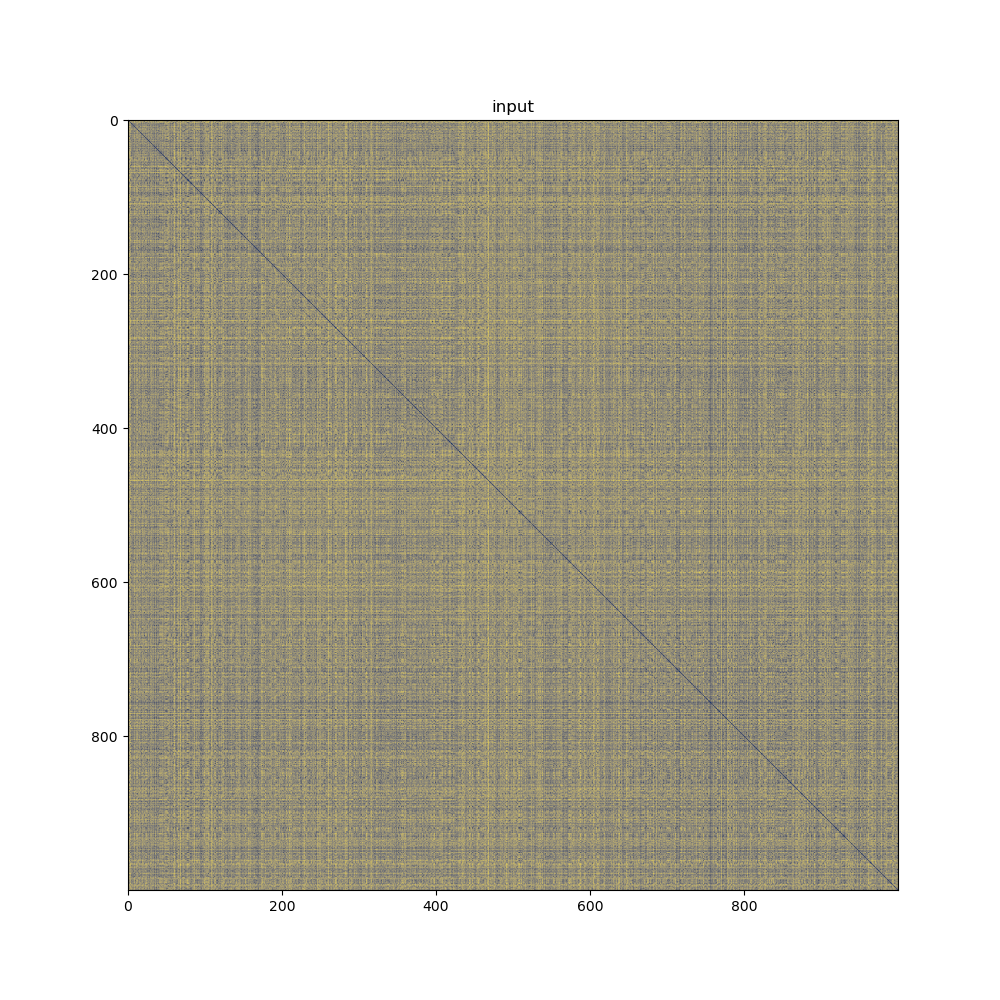}
  \includegraphics[width=0.5\textwidth]{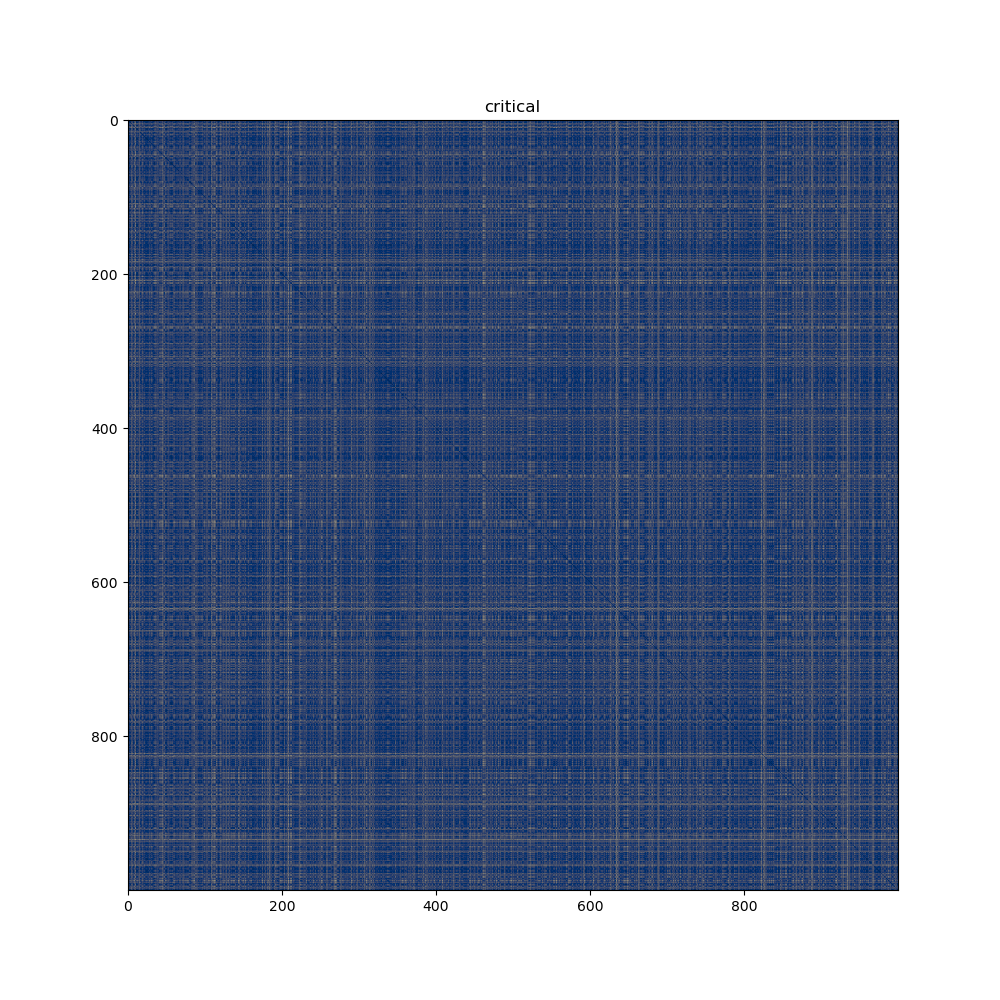}
  \caption{Left: Input correlation matrix,  Right: Output correlation matrix at
    criticality.}
  \label{fig5a}
\end{figure}

% Let us now think again of the correlation matrix, depicted in figures \ref{fig3}
% and \ref{fig4}, as a statistical mechanics system where each element represent a
% physical quantity, say a  magnetic moment, of value $0<s<1$.
% Now take groups of,
% say 25, neighboring magnetic moments, and replace such group by a single
% magnetic moment value equal to the average of the group. Doing this over the
% whole matrix, will effectively re-scale the size of the matrix by a factor of
% five.
%\footnote{This procedure is known in the physical literature as the {\it
%renormalization group flow}, see for example \cite{yeomans}. }
% By keep performing this procedure it can be observed, similarly as in simulated physical second phase transitions, that after few iterations, the order phase will {\it flow} to a complete correlated system, the disorder phase will flow to complete uncorrelated system, but what is more striking, the critical phase will remain approximately invariant, and will not flow to any of the two other extremes.
We have observed in table \ref{tab1} that the input
mean correlation for MNIST data set is not too far from the output mean correlation at
criticality,
which is a consequence of the approximate scaling symmetry in
deepness space. We expect such scaling symmetry to be reflected in
the output correlation matrix, or in other words: since input mean correlations are
approximately preserved for propagation at the critical phase, we should be able
to re-size the input correlation matrix in a way that the input mean
correlation is approximately preserved in the output mean correlation, and {\it our conjecture is that such mean correlation symmetry
  is reflected as a symmetry of learning performance at criticality}. More
concretely, due to the law of large numbers, mean correlations of
sub-correlation matrices approximate the mean correlation of the whole matrix.
We then can take a subsample of the input data, preserving the input mean correlation. Since
such mean correlation is approximately preserved through the network at
criticality, the mean correlation of the output correlation matrix does not
change appreciably and as a consequence the learning map should not change much
and therefore, resizing the input data will preserve (approximately) learning
performance,  as we will explore experimentally in what follows. For that, we compute the normalized correlation matrix and
the mean correlation for the input MNIST data set and propagate the data through
the network, in order to compute the matrix correlation and mean correlation of the propagated output. Next, we reduce the size of the input data and compute once again
mean correlations for input and output.  The graphical representation of input
and output correlation matrices is shown in figure
\ref{fig5a} for 1000 MNIST examples
The corresponding mean correlation for the input $1000\times 1000$ and $500\times 500$
correlation matrices are shown in table \ref{tab2} where we can check the
conservation of output mean correlations from input resizing.
\begin{table}[ht]
  \caption{Mean Correlations for Whole Data Subset and Half Data Subset}
  \begin{center}
    \begin{tabular}{ |c|c|c||c|c| } 
      \hline
      & Input & Half-Input & Critical Output & Half-Input Critical Output \\
      \hline
      $ \langle c\rangle$ & 0.40 & 0.39 & 0.72 & 0.72  \\
      \hline
    \end{tabular}
    \label{tab2} 
  \end{center}
\end{table}

Table \ref{tab2} shows that propagating half the input data has barely any
effect on the mean correlation of the output. In next section we will explore
the impact of this data resizing on training. 

% \begin{figure}[!htb] 
%   \includegraphics[width=0.6\textwidth]{reno_critical_exp0.png}  \includegraphics[width=0.35\textwidth]{keno_critical_exp1.png}  \includegraphics[width=0.15\textwidth]{reno_critical_exp2.png}
%   \caption{Center Upper: Output correlation matrix at criticality, Lower Right: First rescaling output correlation matrix. Lower Left: Second re-scaling output correlation matrix.  }
%   \label{fig5b}
% \end{figure}
% From figure \ref{fig5a} and \ref{fig5b} we can appreciate a very well approximated scaling symmetry for the MNIST input data and a well inherited scaling symmetry for the output layer at the critical phase. We wonder if this inherited scaling will have any impact in learning performance. To try to answer that question we will perform some experiments in the next section.

\section{Experiments}
\subsection{Reducing Input Data}
Based on last section's discussion, we can intuitively think that at critical
initialization, we can reduce the amount of input data without considerably degrading accuracy in the learned input-output map. In this section we will provide experimental evidence that is the case.

We train a very simple feedforward network on the MNIST data set. Our
network consist of 6 hidden layers \footnote{More than two feedforward layers
  to learn the classification problem of MNIST is a bit overkill, but here we
  are interested in the impact of many layers.}  and a 10 units output layer representing the
output probabilities for the digits from 0 to 9 \footnote{In this section  we
  add a 10-units output layer corresponding to the 10 digits of the MNIST
  dataset.}. We use stochastic gradient descent on a cross-entropy cost function. Since we are only worry for the performance of the network as a function of initial data size, we  will not use any particular regularization.

\begin{figure}[!htb]
  \includegraphics[width=0.5\textwidth]{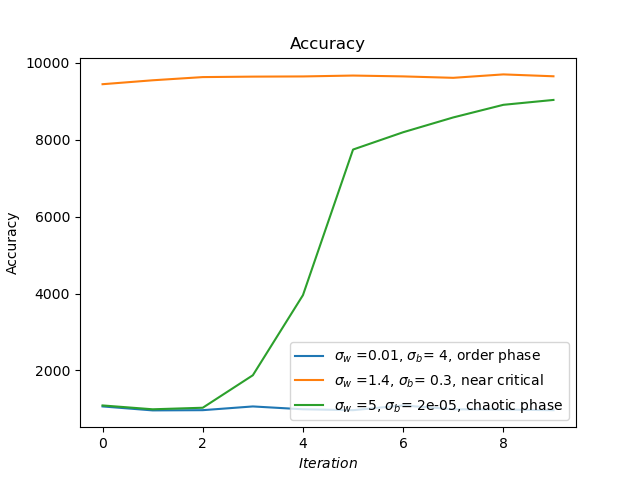}
  \includegraphics[width=0.5\textwidth]{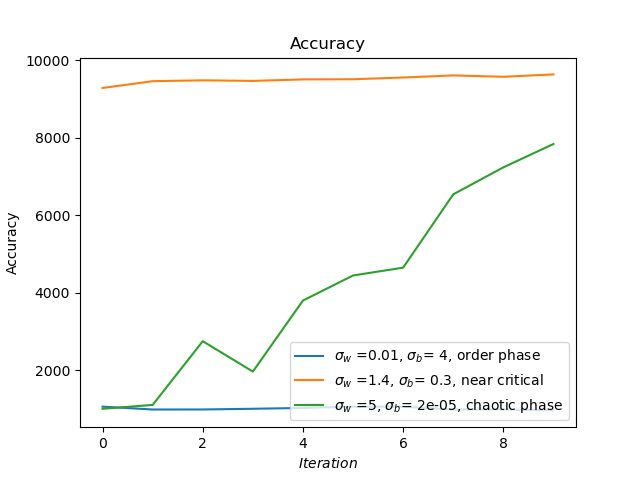} 
  
  \caption{Left: Accuracy for several epochs at the ordered, disordered and
    critical initializations for 50000 input MNIST examples, Right:  Accuracy for several epochs at the ordered, disordered and critical initializations for 25000 input MNIST examples.}
  \label{fig6}
\end{figure}

First we train the network with 50000 MNIST examples, and let it run for just
some few epochs, getting the accuracies over validation (unseen) data in the
three phases as plotted in figure \ref{fig6} left.
Then we re-train the same architecture but with half the examples, i.e 25000,
and from the right figure \ref{fig6} we can observe that, even though the accuracy
and learning speed at the order phase deteriorates considerably (the chaotic
case is already very deteriorated even with the full dataset), the performance of the network at the critical phase does not diminish at all, with the advantage that with that  many less examples, the  network takes  almost half the time in  running the same number of epochs. 

Just out of curiosity, we push harder, and re-train the network with only 15000
examples, and even thought the accuracy at the critical initialization suffer
some deterioration, is not as strong as in the chaotic and order phases where
the accuracy and learning speed do get very badly behaved. For some practical
cases, the small punishment on the accuracy of the learned map at the critical
phase maybe beneficial in terms of the smaller number of required input data examples and decreased training time, which in our experiments is three times smaller than when in use of the full data set.
\begin{figure}[!htb]
  \centering 
  \includegraphics[width=0.5\textwidth]{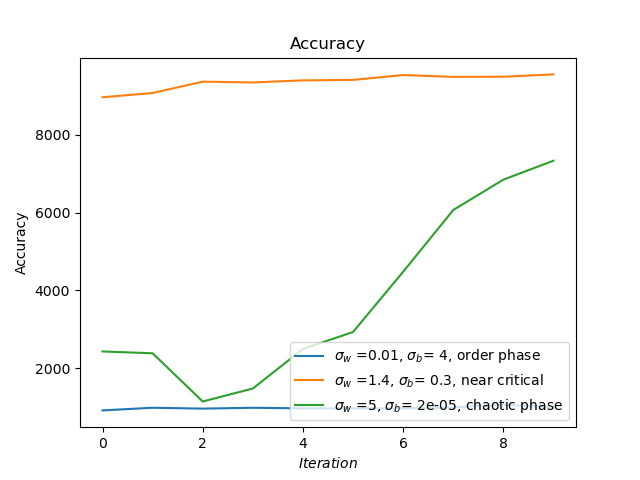} 
  \caption{Accuracy for several epochs at the ordered, disordered and critical initializations for 15000 input MNIST examples.}
  \label{fig7}
\end{figure}

\subsection{Reducing Hidden Width and Batch Size}
Similarly as to the intuition that resizing of the input data is an approximate
symmetry of the output correlation matrix at criticality, we can consider the effect of resizing the width of the hidden layers, in this section we take the same architecture as in the previous subsection but will resize the hidden layers by half, however preserving the same size of the input data.

\begin{figure}[!htb]
  \centering 
  \includegraphics[width=0.5\textwidth]{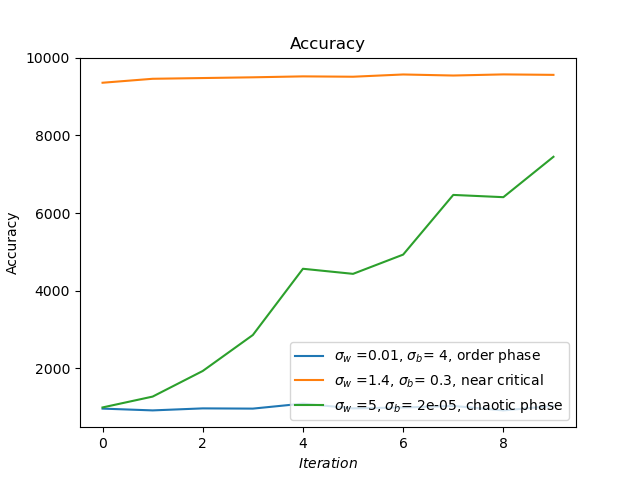} 
  \caption{Accuracy for several epochs at the ordered, disordered and critical initializations with half-reduced hidden width.}
  \label{fig8}
\end{figure}

As in previous section, reducing the hidden width, has almost not detrimental
impact on the performance of the network initialized at the critical phase,
while it does hurt considerably the performance at initializations in the order
and disorder phases, as shown in figure \ref{fig8}.

\begin{figure}[!htb]
  \centering 
  \includegraphics[width=0.5\textwidth]{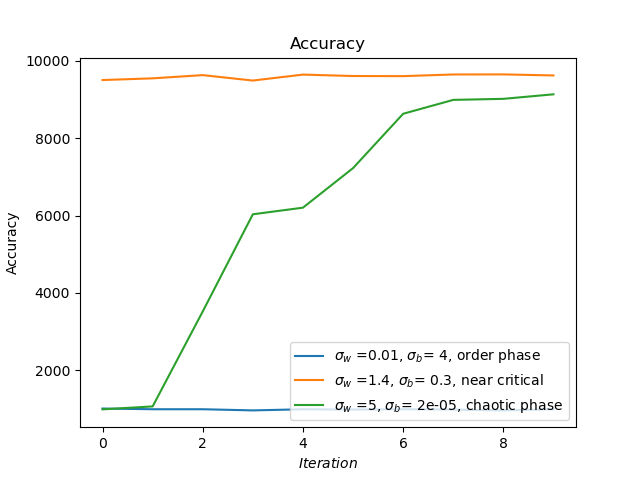}
  \caption{Accuracy for several epochs at the ordered, disordered and critical initializations with half-reduced batch size for stochastic gradient descent.}
  \label{fig9}
\end{figure}

Lastly, we can also examine the impact of resizing the batch for
stochastic gradient descent, while keeping the original sizes for the input data
and hidden layers width. We reduce the size of this batch by a half and plot the
resulting accuracies over the validation set in figure \ref{fig9}, once again
observing little impact performance on initialization at the critical phase,
unlike the other two phases.

\section{Discussion and Conclusions}
In this paper we have studied the phase transition occurring in the propagation
of the covariance matrix through a random deep feedforward network. In summary,
we have shown that a naive statistical physics description for such system can
be developed, which in turns lead us to consider some well known properties of physical
phase transitions, such as scaling symmetry at criticality. We additionally
argued, that the given scaling symmetry translates into a data-resizing symmetry
which in principle, would allow to scale down a large network into a smaller
one without degrading learning performance.

It is worth mentioning that more precise correspondences between neural networks and
statistical physical system have been developed previously in several other works. Such as those recently
developed at \cite{schoenholz2017correspondence} and \cite{choromanska2015loss}.
It would be very interesting to studied the phase transition treated in this
note, from the point of view of such interesting physical approaches.

Another recent interesting relation between physical phase transition and deep neural
networks have been studied in the context of the loss landscape in \cite{Franz_2016, Geiger_2019, ziyin2022exact}. It would be interesting to ask if the phase
transitions in random networks studied in this paper has any relation to those
studied there.
\newpage
%%%%%%%%%%%%%%%%%%%%%%%%%%%%%%%%%%%%%%%%%%%%
%%%%%%%%%%%commented out%%%%%%%%%%%%%%%%%
\bibliographystyle{unsrt}
\bibliography{phase_tran_NN}
\end{document}